\documentclass{article}
\usepackage{spconf,amsmath}
\usepackage{graphicx} \graphicspath{ {figures/} }
\usepackage{multicol, multirow, booktabs, microtype}

\usepackage{balance}
\usepackage[pdftex, pagebackref=true, pdfpagelabels=true,
hypertexnames=true, plainpages=false, naturalnames=false,
bookmarks=true, bookmarksnumbered=true,
pdfpagemode=UseOutlines,]{hyperref}

\hypersetup{colorlinks, citecolor=[rgb]{0,0.6,0}, filecolor=[rgb]{0.6,0,0}, linkcolor=[rgb]{0.6,0,0}, urlcolor=[rgb]{0.6,0,0.6}}
\title{MultiMix: Sparingly Supervised, Extreme Multitask Learning\\ From Medical Images}
\name{Ayaan Haque$^{1\star}$,
\thanks{$^{\star}$Authors contributed equally} Abdullah-Al-Zubaer Imran$^{2,3\star}$,\footnotemark[1]
Adam Wang$^{2}$,
Demetri Terzopoulos$^{3,4}$}
\address{ 
$^{1}$Saratoga High School, Saratoga, CA, USA \\
$^{2}$Stanford University, Stanford, CA, USA\\
$^{3}$University of California, Los Angeles, CA, USA\\
$^{4}$VoxelCloud, Inc., Los Angeles, CA, USA
}

\begin{document}
\maketitle
\begin{abstract}
Semi-supervised learning via learning from limited quantities of labeled data has been investigated as an alternative to supervised counterparts. Maximizing knowledge gains from copious unlabeled data benefit semi-supervised learning settings. Moreover, learning multiple tasks within the same model further improves model generalizability. We propose a novel multitask learning model, namely MultiMix, which jointly learns disease classification and anatomical segmentation in a sparingly supervised manner, while preserving explainability through bridge saliency between the two tasks. Our extensive experimentation with varied quantities of labeled data in the training sets justify the effectiveness of our multitasking model for the classification of pneumonia and segmentation of lungs from chest X-ray images. Moreover, both in-domain and cross-domain evaluations across the tasks further showcase the potential of our model to adapt to challenging generalization scenarios.\footnote{Code, pretrained models, and additional details are available at \href{https://github.com/ayaanzhaque/MultiMix}{https://github.com/ayaanzhaque/MultiMix}}
\end{abstract}


\section{Introduction}
\label{sec:intro}

Learning-based medical image analysis has become widespread with the advent of deep learning, especially with Convolutional Neural Networks (CNNs). However, deep learning models are mostly reliant on large pools of labeled data. Especially in the medical domain, obtaining labeled images is often infeasible, as annotation requires domain expertise and manual labor, making it difficult to train large-scale deep learning models. To address the limited labeled data problem in image analysis tasks, Semi-Supervised Learning (SSL) has been gaining attention. In SSL, unlabeled examples are leveraged in combination with labeled examples to maximize information gains \cite{chapelle2009semi}. Furthermore, MultiTask Learning (MTL) has been researched for improving the generalizability of any models \cite{ruder2017overview}. MTL is defined as optimizing more than one loss in a single model such that multiple related tasks are performed through shared representation learning.

\label{sec:method}
\begin{figure}
    \centering
    \includegraphics[width=\linewidth]{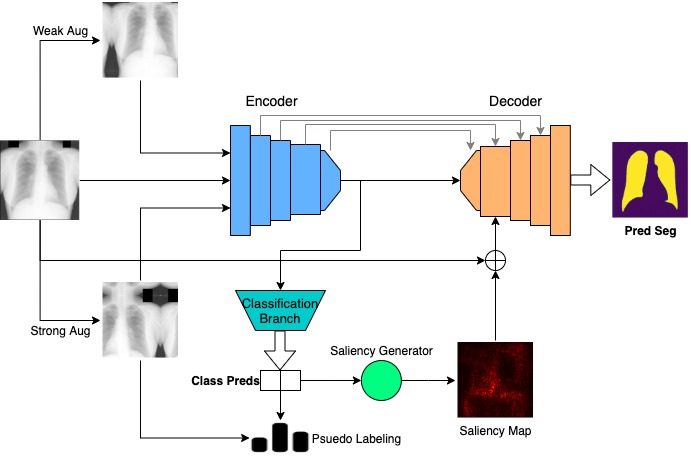}
    \caption{Schematic of the MultiMix model: (classification) Using predictions on unlabeled weakly augmented images, pseudo-labels are generated with confidence, and loss is computed with these labels and the strongly augmented versions of those images. (segmentation) Generated saliency maps from the class predictions are concatenated via the saliency bridge module to guide the decoder for the final segmentations.}
    \label{fig:model}
\end{figure}

Semi-supervised multitask learning is still under-explored in medical imaging. Some work in this direction has recently been published \cite{imran2019semi, liu2008semi, mehta2018ynet}. 
A saliency map enables the analysis of model predictions through the visualization of meaningful visual features \cite{simonyan2014deep}. While any deep learning model can be investigated for better explainability via the saliency map, to our knowledge, bridge saliency between the two shared tasks in a single model has not yet been explored in the literature. We show that combining a bridge-saliency module and a simple yet effective semi-supervised learning method in a multitasking setting can yield improved and consistent performance across multiple domains. 

Our main contributions may be summarized as follows:
\begin{itemize}
\item 
A novel semi-supervised method exploiting consistency augmentation and multi-source data for jointly learning diagnostic classification and anatomical structure segmentation.

\item 
An innovative saliency bridge module connecting the two related tasks within the same model.

\item 
Extensive experimentation with varied quantities of labeled data from different sources (both in- and cross-domain) demonstrating the improved generalizability of our model.
\end{itemize}

\section{MultiMix}

To formulate the problem, we assume unknown data distributions $p(X^s,Y)$ over images $X^s$ and segmentation labels $Y$, and $p(X^c, C)$ over images $X^c$ and class labels $C$. We also assume access to labeled training sets $\mathcal{D}^s_l$ sampled i.i.d.~from $p(X^s,Y)$ and $\mathcal{D}^c_l$ sampled i.i.d.~from $p(X^c, C)$, as well as unlabeled training sets $\mathcal{D}^s_u$ sampled i.i.d.~from $p(X^s)$ and  $\mathcal{D}^c_u$ sampled i.i.d.~from $p(X^c)$ after marginalizing out $Y$ and $C$, respectively. In the proposed MultiMix model (Fig.~\ref{fig:model}), we utilize a U-Net-like encoder-decoder architecture \cite{ronneberger2015unet} for image deconstruction and reconstruction. The encoder functions similarly to a standard convolutional neural network. Using pooling layers followed by fully-connected layers, the encoder outputs a classification prediction through the classification branch.

For sparingly-supervised classification, we leverage data augmentation and pseudo-labeling. Inspired by \cite{sohn2020fixmatch}, we take an unlabeled image and perform two separate augmentations. A single unlabeled image is first weakly augmented, and from that weakly augmented version of the image, a pseudo-label is assumed based on the prediction from the current state of the model. Secondly, the same unlabeled image is then augmented strongly, and a loss is calculated with the pseudo-label from the weakly augmented image and the strongly augmented image itself. Note that this image-label pair is retained only if the confidence with which the model generates the pseudo-label is above a tuned threshold, which prevents the model from learning from incorrect and poor labels. Weak augmentations consist of standard augmentations and are applied to labeled data as well, while strongly augmented images are augmented by randomly applying heavy augmentations from a pool of augmentations.

The classification loss is therefore calculated as
\begin{equation}
\label{eqn:class-loss1}
       L^c = L_{l}(\hat{c}_l, c_l) + \lambda L_{u}(\hat{c}_s, arg\max(\hat{c}_w) \ge t),
\end{equation}
where $L_l$ is the supervised loss, which uses cross-entropy, $\hat{c}_l$ denotes the predictions for input $x^c_l$ with the corresponding reference labels $c_l$ and $\hat{c}_s$ are the classification predictions on strongly augmented images,  parameter $\lambda$ is the unsupervised classification loss weight, $L_u$ is the unsupervised loss, and $arg\max(\hat{c}_w) \ge t$ is the pseudo-labeling function, where $\hat{c}_w$ are the predictions on the weakly augmented images and $t$ is the pseudo-labeling threshold.

We generate saliency maps based on the predicted classes using the gradients of the encoder. While the segmentation images do not necessarily represent pneumonia, the classification task, the generated maps highlight the lungs, creating images at the final segmentation resolution. We hypothesize that these saliency maps can be used to guide the segmentation during the decoder phase, yielding improved segmentation while learning from limited labeled data. In our algorithm, the generated saliency maps are concatenated with the input images, downsampled, and added to the feature maps input to the first decoder stage. Moreover, to ensure consistency, we compute the KL divergence between segmentation predictions for labeled and unlabeled examples. This penalizes the model from making predictions that are increasingly different than those of the labeled data, which helps the model fit more appropriately for the unlabeled data.

The segmentation loss is therefore:
\begin{equation}
\label{eqn:class-loss2}
       L^s =  \alpha L_{l}(\hat{y}_l, y_l) + \beta L_{u}(\hat{y}_l, \hat{y}_u),
\end{equation}
where $\alpha$ is the segmentation loss weight, which uses dice loss, $\hat{y}_l$ are the labeled segmentation image predictions, $y_l$ are the corresponding labels, $\beta$ is the unsupervised segmentation loss weight, and $\hat{y}_u$ are the unlabeled segmentation image predictions.

\section{Experimental Evaluation}
\label{sec:exp}

%%%%%%%%%%%%%%%%%%%%%%%%%%%%%%%%%%%%%%%%%%%%%%%%
%Table: Data splits 
\begin{table}
\setlength{\tabcolsep}{4pt}
\centering
\caption{Details of the datasets used for training and testing.}
\medskip
\label{table:data}
\resizebox{\linewidth}{!}{
\begin{tabular}{@{} cc lc ccc c ccc @{}}
\toprule
mode & \phantom{a} & Dataset & \phantom{a} & Total & Normal & Abnormal & \phantom{a} & Train & Val & Test\\
\midrule
\multirow{2}{*}{in-domain} && JSRT && 247 & -- & -- && 111 & 13 & 123\\
%\cmidrule{3-11}
&& CheX && 5,856 & 4273 & 1583 && 5216 & 16 & 624\\
\midrule
\multirow{2}{*}{cross-domain} && MCU && 138 & -- & -- && 93 & 10 & 35\\
%\cmidrule{3-11}
&& NIHX && 4185 & 2754 & 1431 && -- & -- & 4185\\
\bottomrule
\end{tabular}
}
\end{table}
%%%%%%%%%%%%%%%%%%%%%%%%%%%%%%%%%%%%%%%%%%%%%%%%%%

\begin{figure*}
    \centering
    \resizebox{\linewidth}{!}{
    \includegraphics[width=0.49\linewidth]{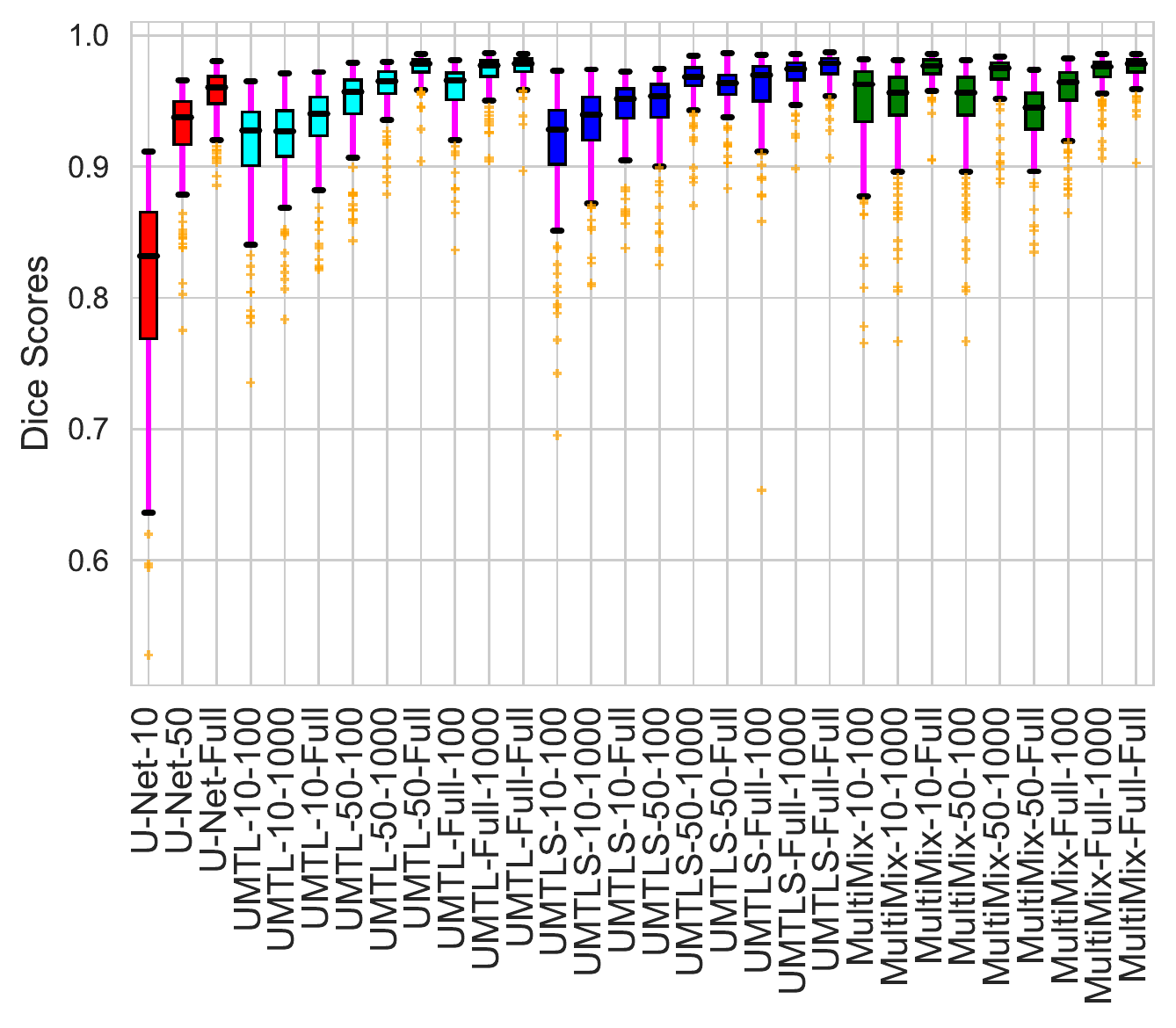}
    \hfill
    \includegraphics[width=0.49\linewidth]{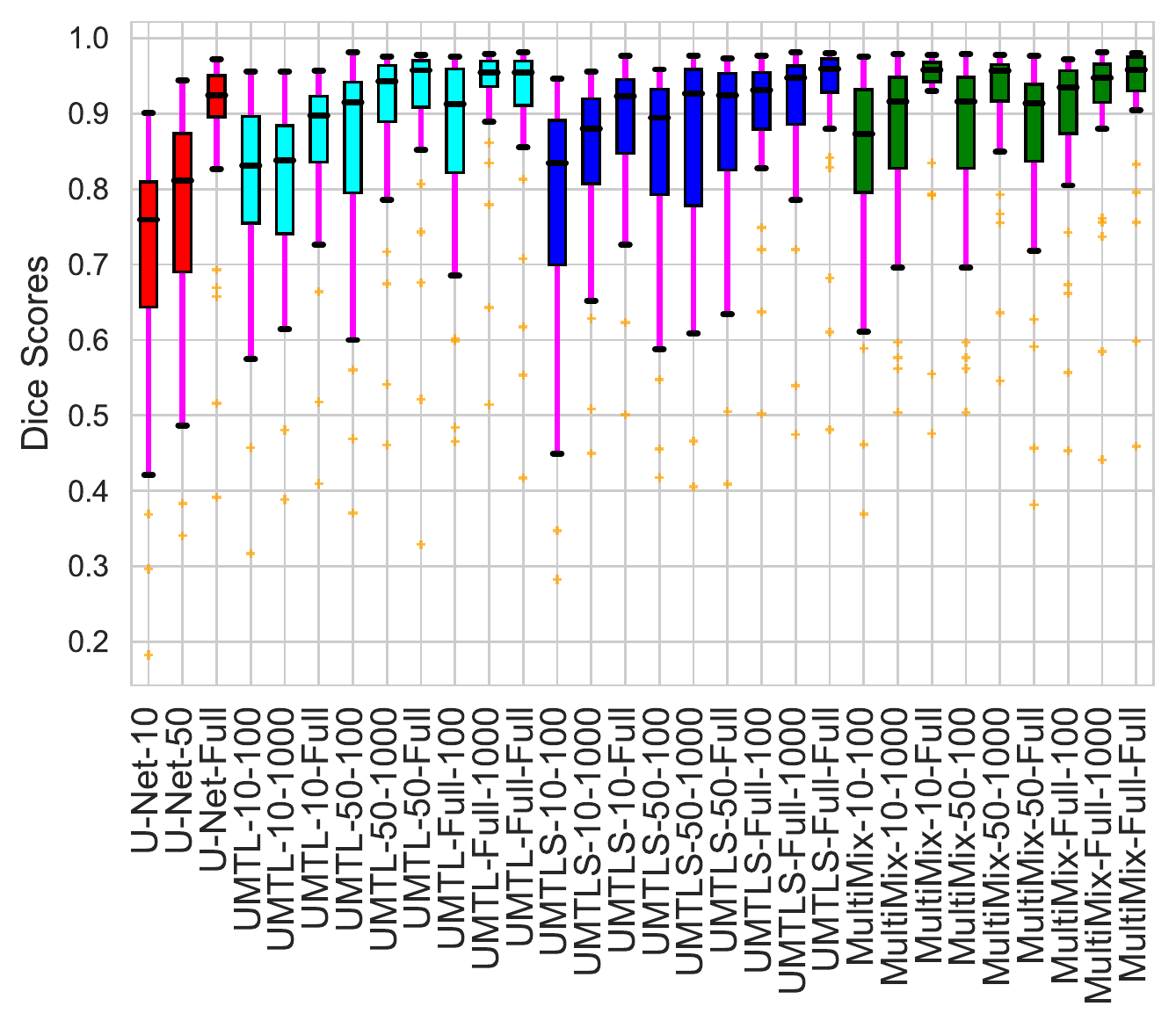}
    }
    \caption{Box plots of Dice scores from baselines and MultiMix: in-domain (left) and cross-domain (right).}%, and naming is model-$|\mathcal{D}^c_l|$-$|\mathcal{D}^s_l|$.}
    \label{fig:consistency}
\end{figure*}

\subsection{Implementation Details}

\textbf{Data:} The models were trained and tested for the combined classification and segmentation tasks using data from two different sources: pneumonia detection (CheX) \cite{kermany2018identifying} and JSRT \cite{shiraishi2000development}. Furthermore, we used the Montgomery County chest X-rays (MCU) \cite{jaeger2014two} and a subset of the NIH chest X-ray dataset (NIHX) \cite{wang2017chestx} for cross-domain evaluation. Table~\ref{table:data} presents the details of the datasets used in our experiments.

\textbf{Inputs:} 
All the images were normalized and resized to $256\times256\times1$ before feeding them to the models. \textbf{Model Architecture:} As the segmentation backbone, we used a U-Net-like encoder-decoder network with skip connections \cite{ronneberger2015unet}, and branched out by adding two pooling layers and an FC layer for the classification predictions. 
We implemented the models using Python with the PyTorch framework using a NVIDIA K80 GPU. \textbf{Baselines:} As baselines, we used the U-Net and encoder-only (Enc) networks separately for the single-task models both in supervised and semi-supervised schemes. Using the same backbone network, we also trained a multitasking U-Net with the described classification branch (UMTL). We incorporated an INorm, LReLU, and dropout at every convolutional block. Moreover, we performed ablation experiments to observe the impact of each of the key pieces in our proposed model: single-task EncSSL (encoder with the SSL algorithm), UMTLS(UMTL with saliency bridge). \textbf{Training:} All the models (single-task or multitask) were trained on varying $|\mathcal{D}^s_l|$ (10, 50, full), and $|\mathcal{D}^c_l|$ (100, 1000, full). Each experiment was repeated 5 times. \textbf{Hyper-parameters:} We used the Adam optimizer with adaptive learning rates of 0.1 every 8 epochs and an initial learning rate of 0.0001. A negative slope of 0.2 is applied to LReLU, and dropout is set to 0.25. We set $t=0.7$, $\lambda=0.25$, $\alpha=5.0$ (for smaller $|\mathcal{D}^s_l|$) and $\beta=0.01$. Each model was trained with a mini-batch size of 10. \textbf{Evaluation (in-domain and cross-domain):} For classification, along with the overall accuracy (Acc), we recorded the class-wise F1 scores (F1-N for normal and F1-P for pneumonia). To evaluate the segmentation performances, the Dice similarity (DS), average Hausdorff distance (HD), and structural similarity measure (SSIM) scores were used.

%%%%%%%%%%%%%%%%Figure: Chest Lung Mask Visualizations (In Domain)%%%%%%%%%%%%%%%%%%%%%
\begin{figure}[t]
\centering
 \resizebox{\linewidth}{!}{%
  \begin{tabular}{l ccc ccc}
  \toprule
   & \multicolumn{3}{c}{\Large JSRT (in-domain)} & \multicolumn{3}{c}{\Large MCU (cross-domain)}\\
   \cmidrule(lr){2-4} \cmidrule(lr){5-7}
    \raisebox{6mm}{\Large GT}
    &
     \includegraphics[width=0.2\linewidth, trim={4cm 1cm 3cm 1cm},clip]{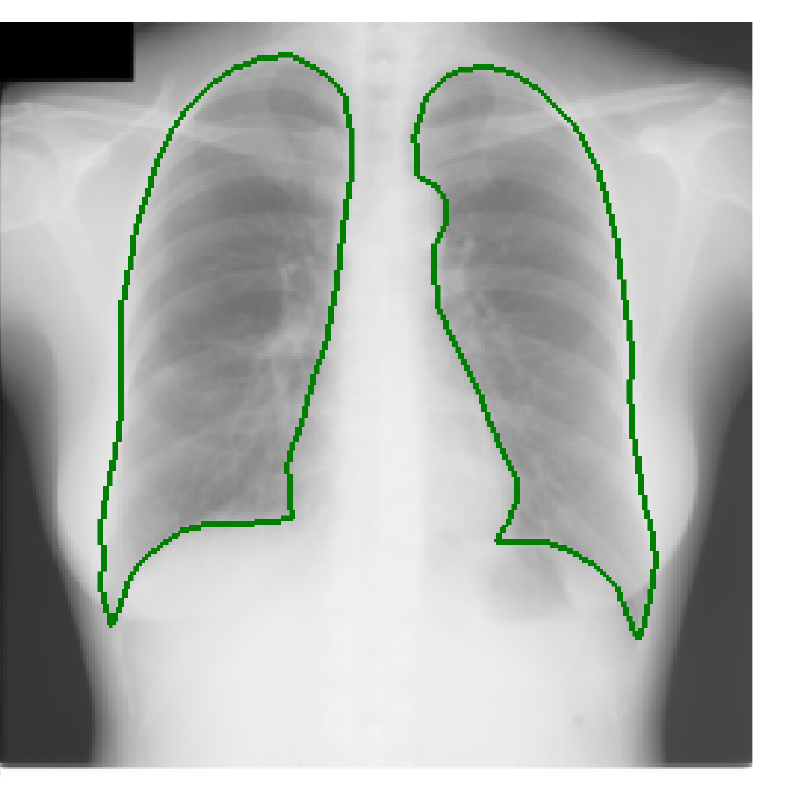}
    &&&
     \includegraphics[width=0.2\linewidth, trim={4cm 1cm 3cm 1cm},clip]{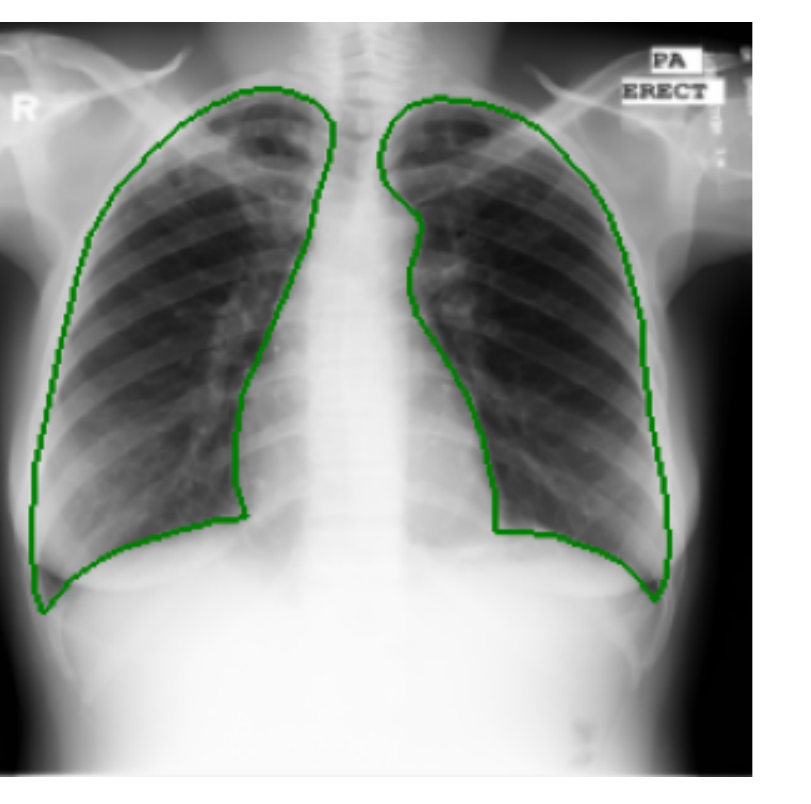}
    \\
    \noalign{\smallskip}
    {\Large $|\mathcal{D}^s|$}
    &
    {\Large 10} & {\Large 50} & {\Large Full} & {\Large 10} & {\Large 50} & {\Large Full}\\

    {\Large\rotatebox{90}{U-Net}}
    &
    \includegraphics[width=0.2\linewidth, trim={4cm 1cm 3cm 1cm},clip]{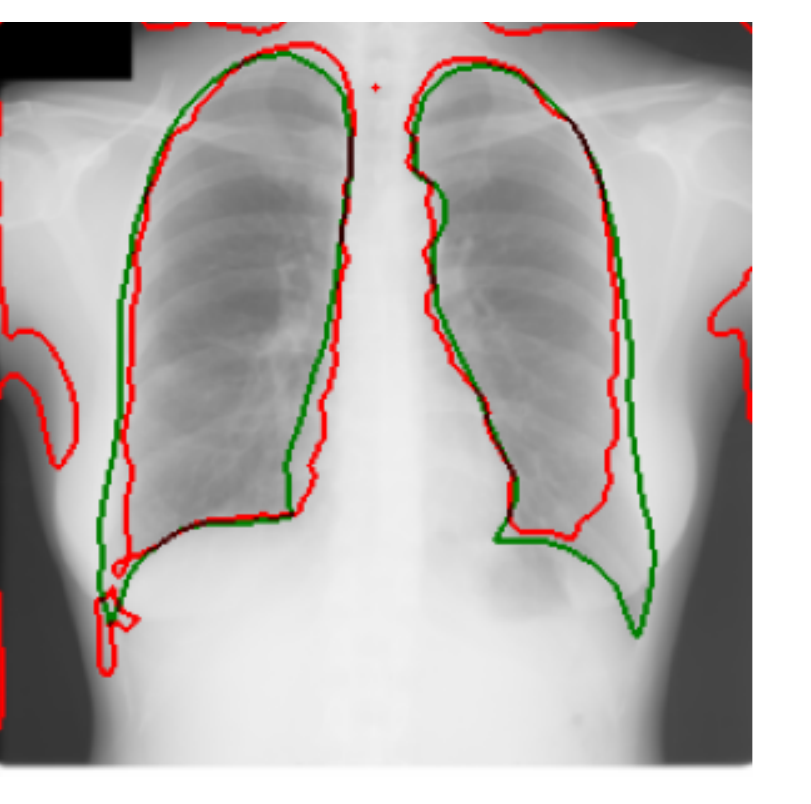}
    &
    \includegraphics[width=0.2\linewidth, trim={4cm 1cm 3cm 1cm},clip]{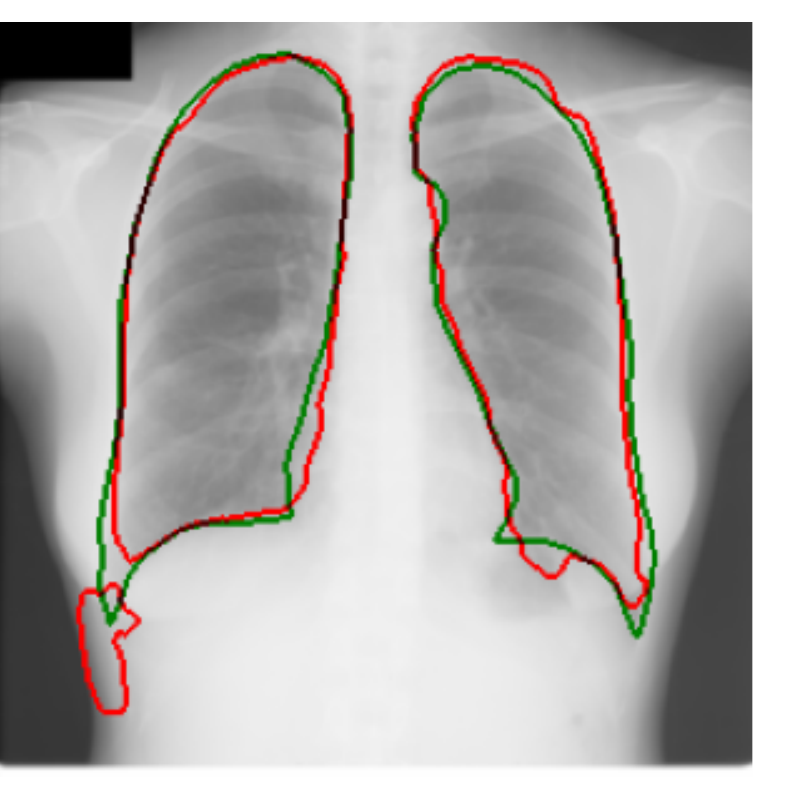}
    &
    \includegraphics[width=0.2\linewidth, trim={4cm 1cm 3cm 1cm},clip]{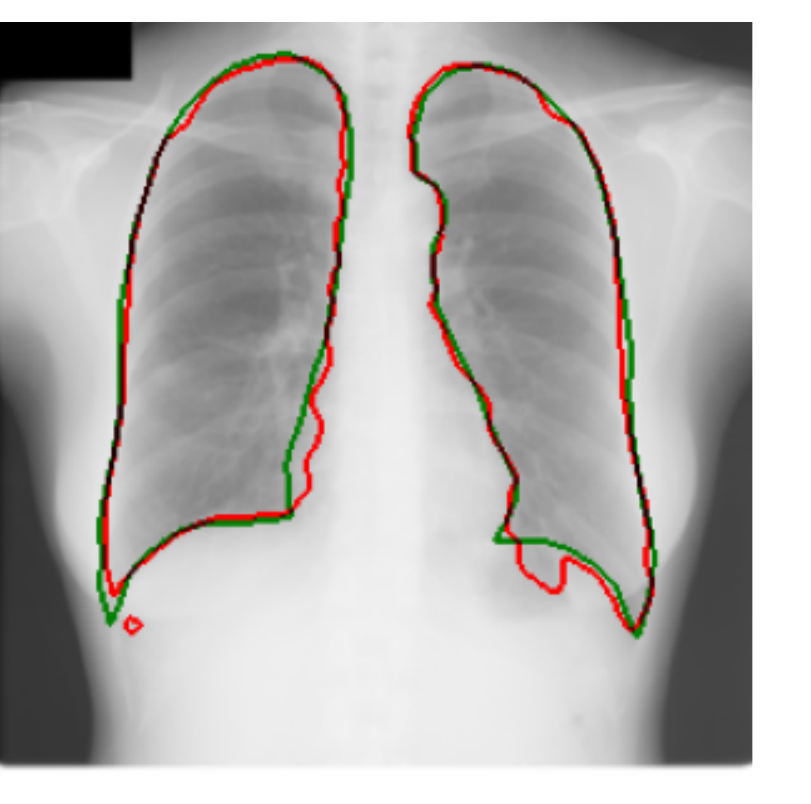}
    &
    \includegraphics[width=0.2\linewidth, trim={4cm 1cm 3cm 1cm},clip]{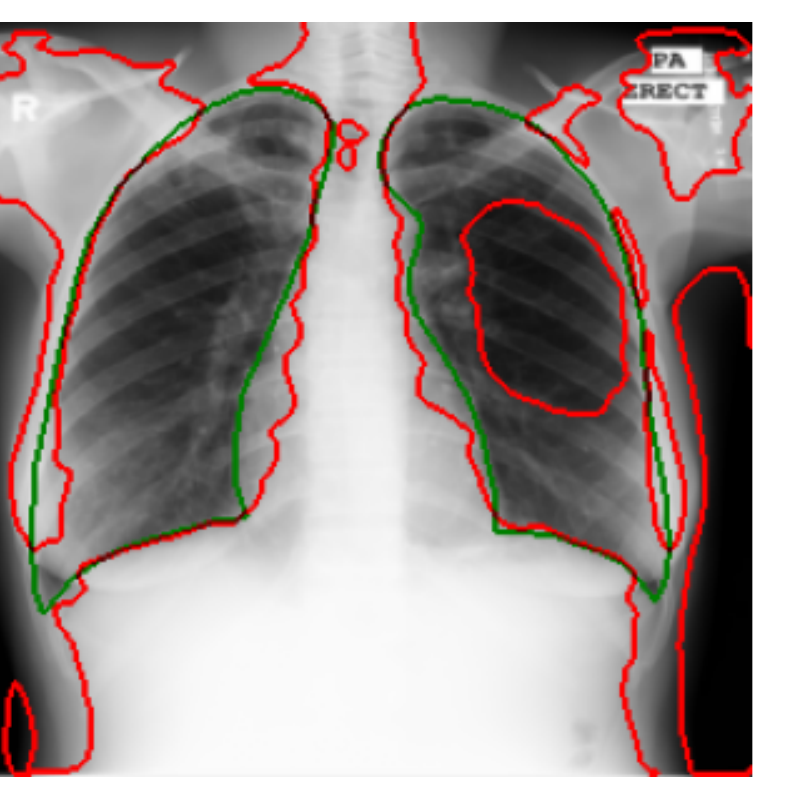}
    &
    \includegraphics[width=0.2\linewidth, trim={4cm 1cm 3cm  1cm},clip]{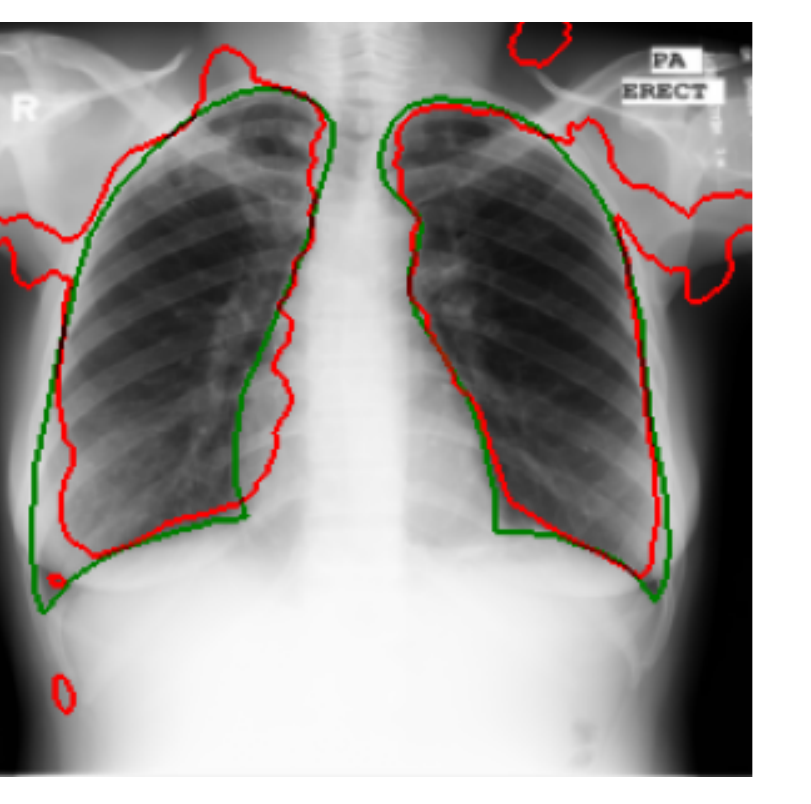}
    &
    \includegraphics[width=0.2\linewidth, trim={4cm 1cm 3cm 1cm},clip]{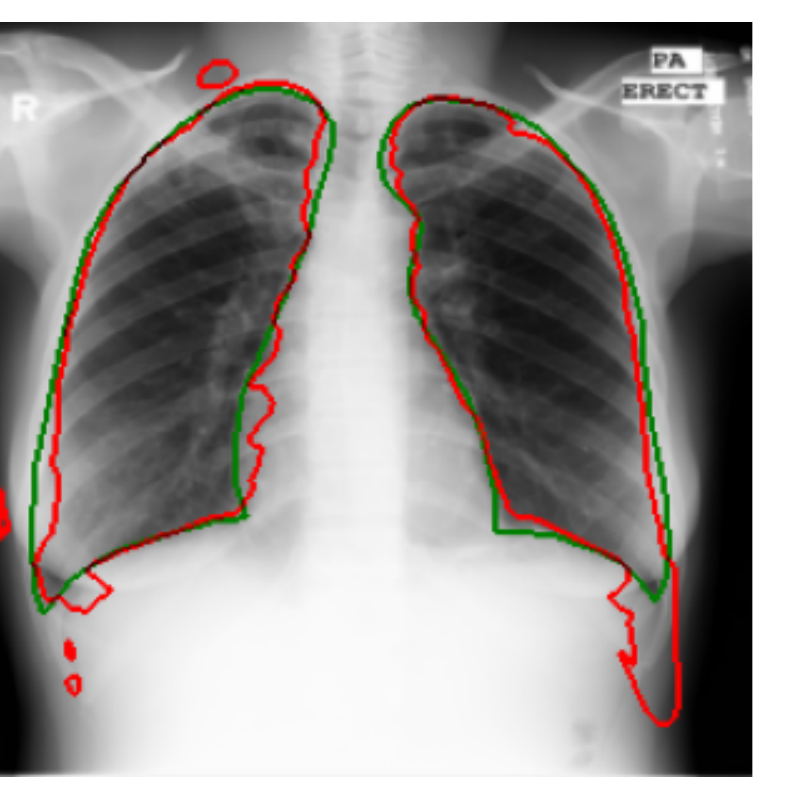}
    \\
    \noalign{\smallskip}
    {\Large\rotatebox{90}{UMTL}} 
    &
    \includegraphics[width=0.2\linewidth, trim={4cm 1cm 3cm 1cm},clip]{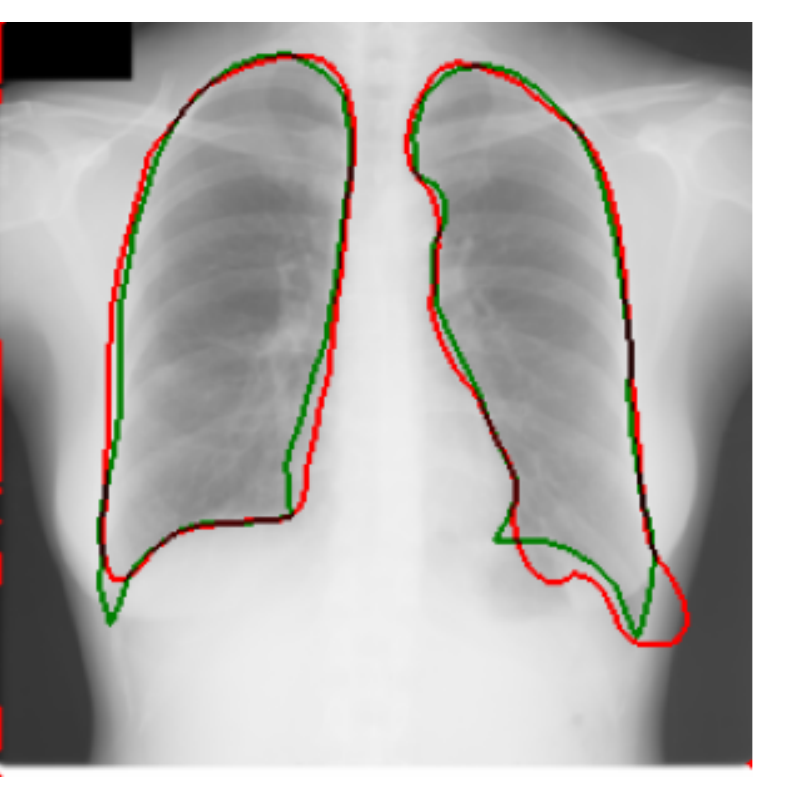}
    &
    \includegraphics[width=0.2\linewidth, trim={4cm 1cm 3cm 1cm},clip]{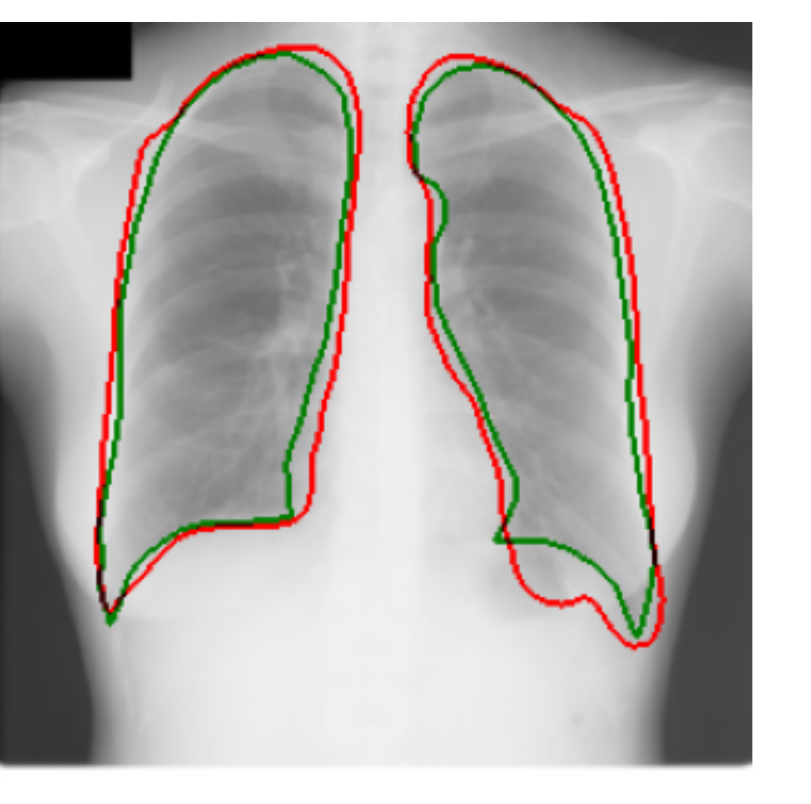}
    &
    \includegraphics[width=0.2\linewidth, trim={4cm 1cm 3cm 1cm},clip]{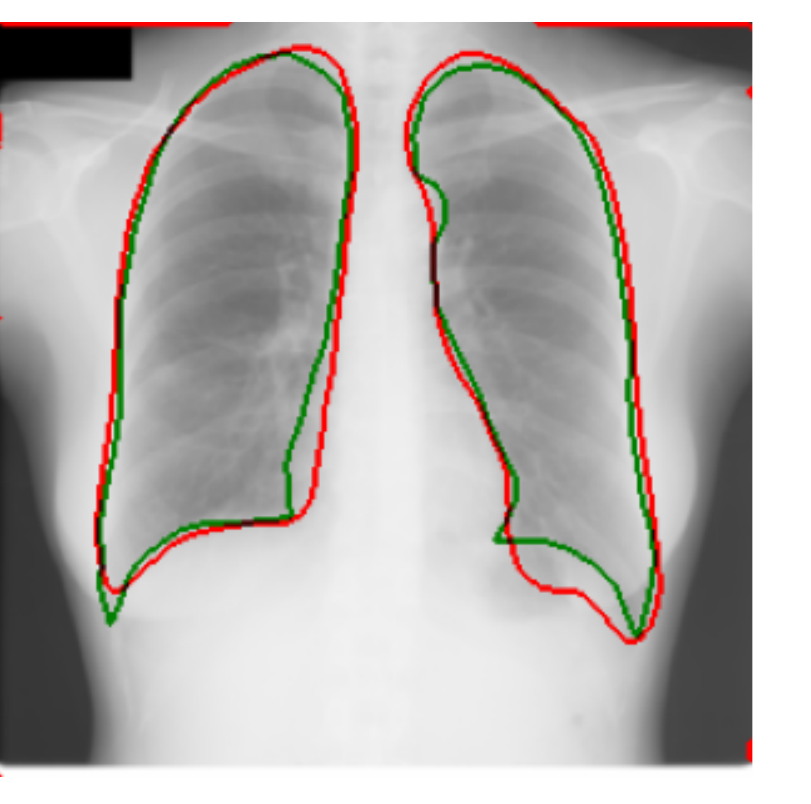}
    &
    \includegraphics[width=0.2\linewidth, trim={4cm 1cm 3cm 1cm},clip]{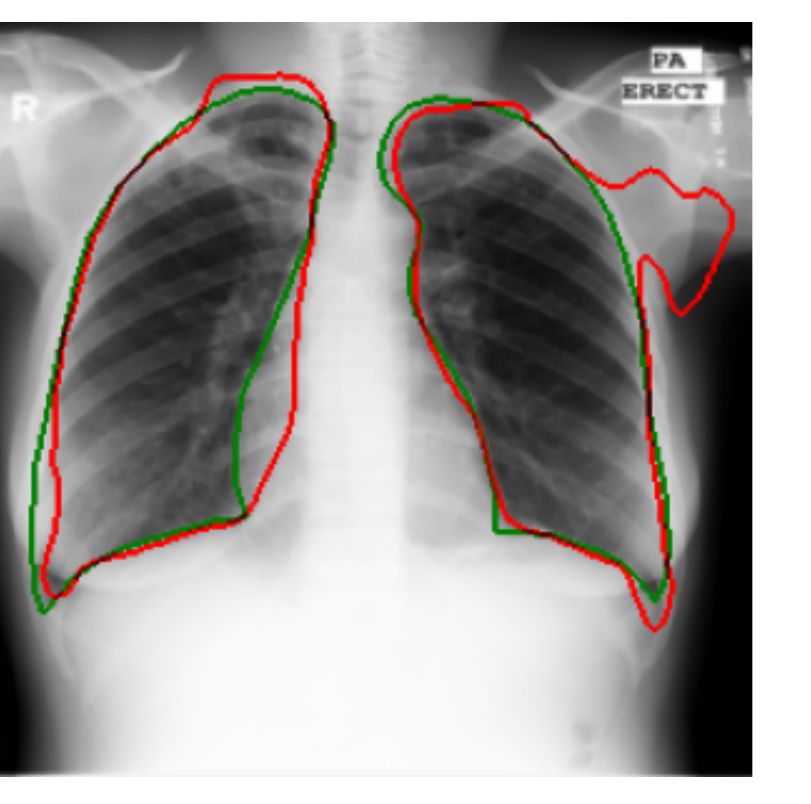}
    &
    \includegraphics[width=0.2\linewidth, trim={4cm 1cm 3cm 1cm},clip]{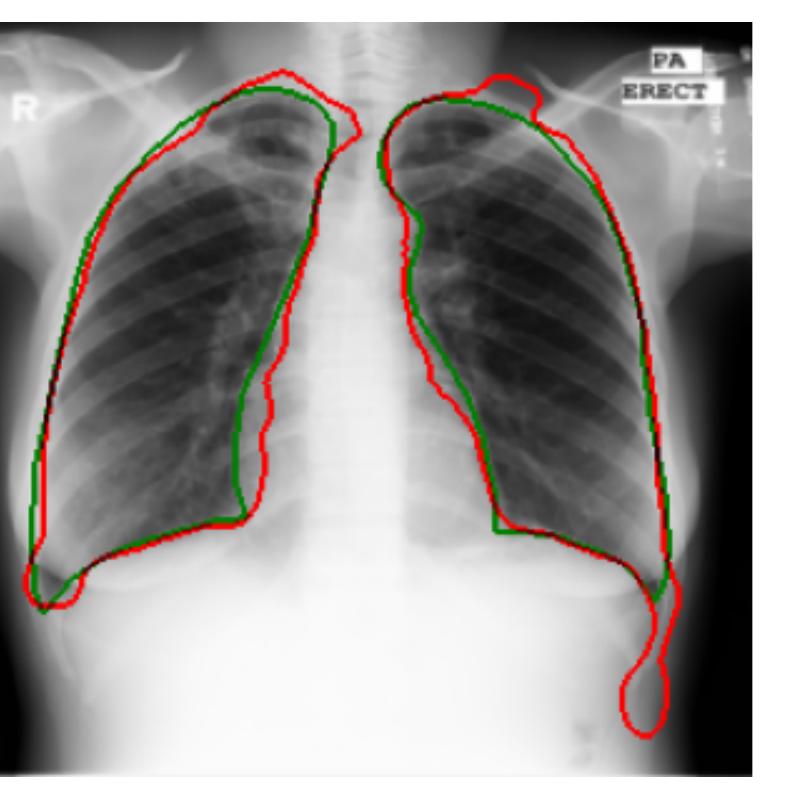}
    &
    \includegraphics[width=0.2\linewidth, trim={4cm 1cm 3cm 1cm},clip]{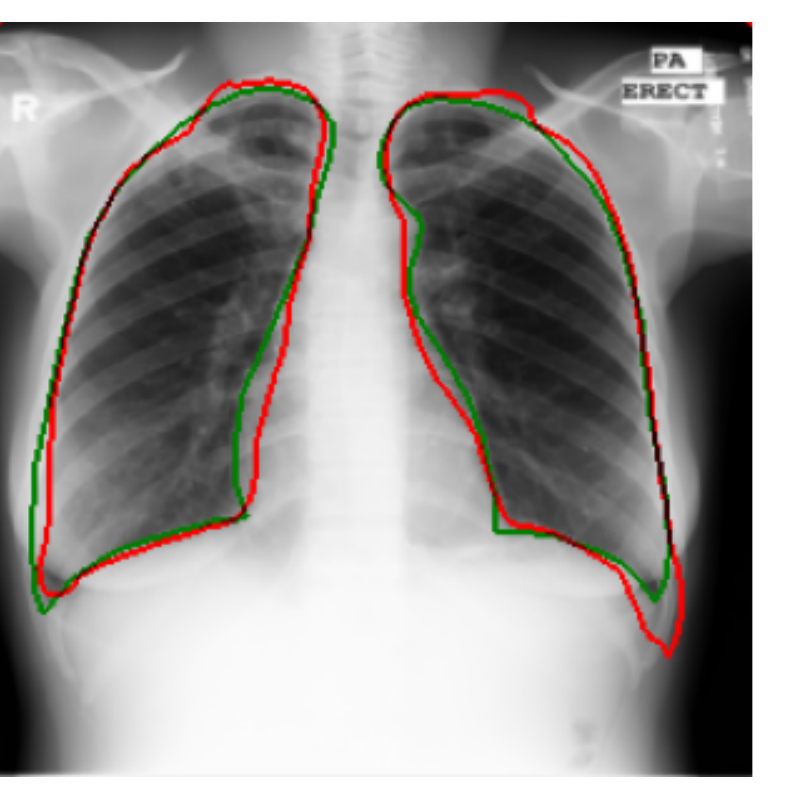}
    \\
    \noalign{\smallskip}
    {\Large\rotatebox{90}{UMTLS}}
    &
    \includegraphics[width=0.2\linewidth, trim={4cm 1cm 3cm 1cm},clip]{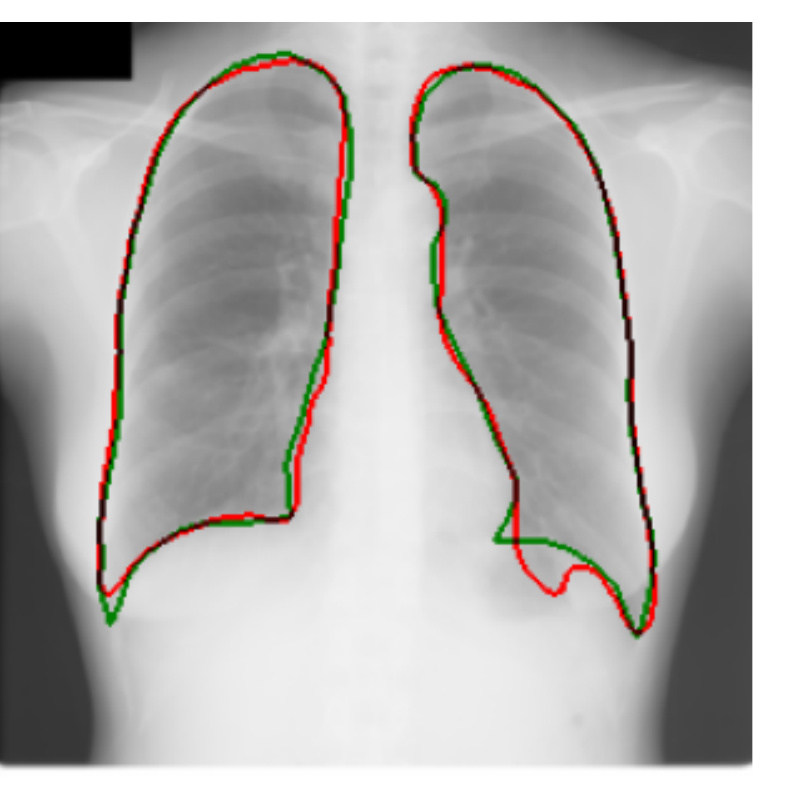}
    &
    \includegraphics[width=0.2\linewidth, trim={4cm 1cm 3cm 1cm},clip]{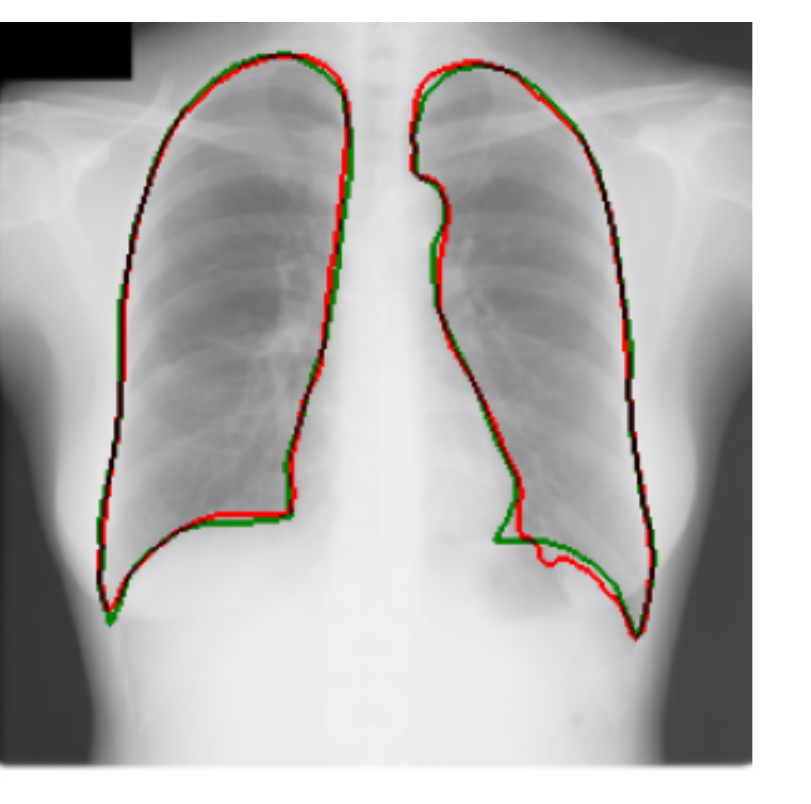}
    &
    \includegraphics[width=0.2\linewidth, trim={4cm 1cm 3cm 1cm},clip]{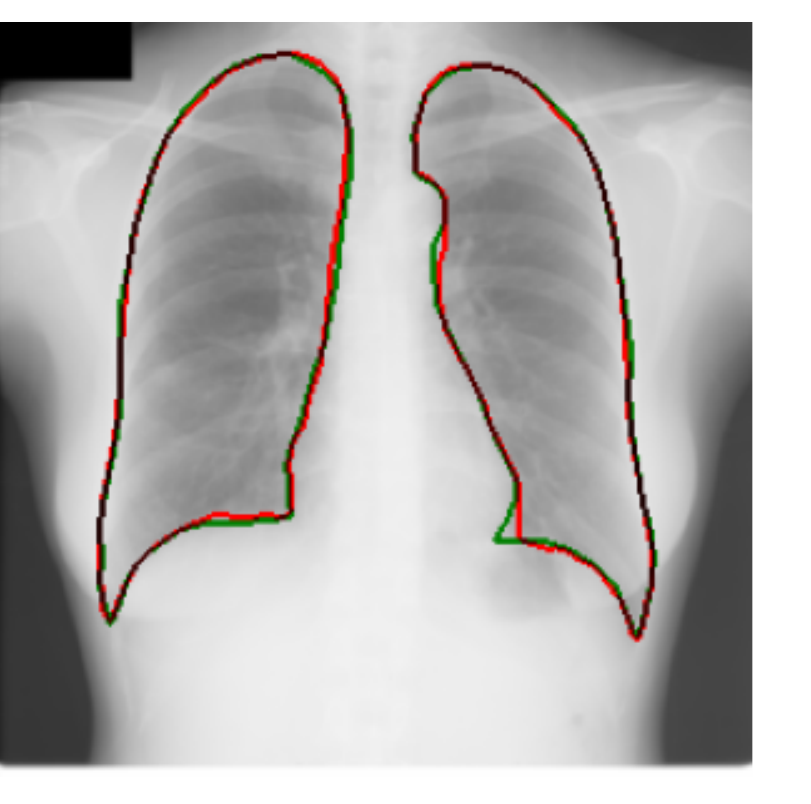}
    &
    \includegraphics[width=0.2\linewidth, trim={4cm 1cm 3cm 1cm},clip]{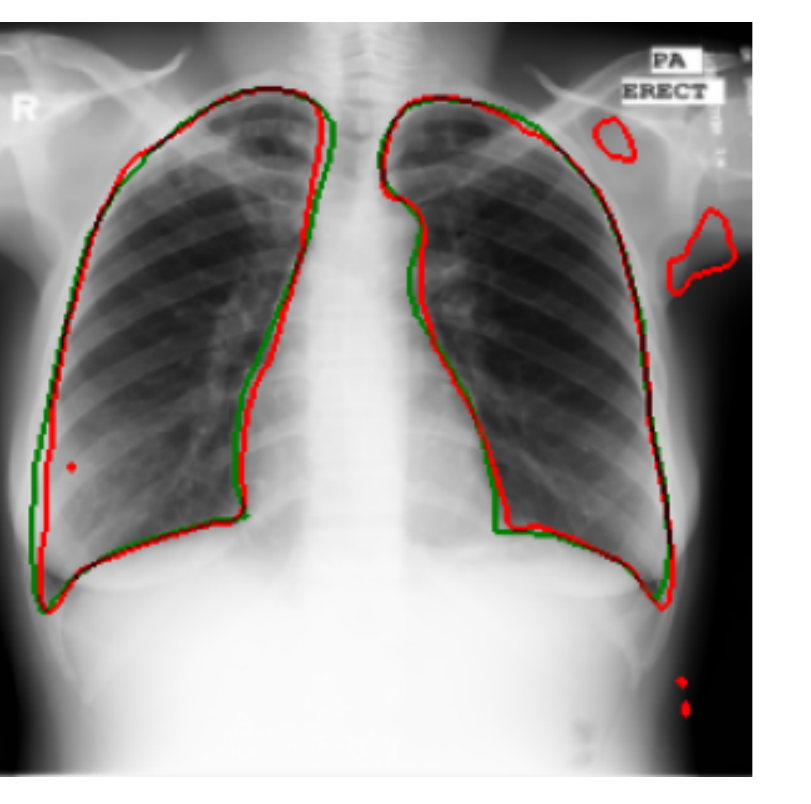}
    &
    \includegraphics[width=0.2\linewidth, trim={4cm 1cm 3cm 1cm},clip]{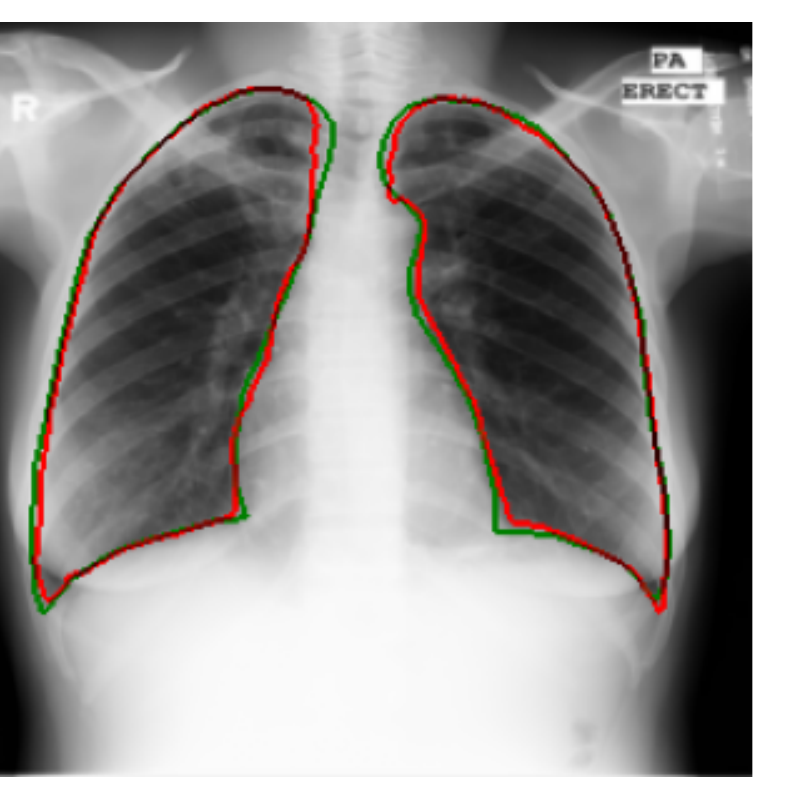}
    &
    \includegraphics[width=0.2\linewidth, trim={4cm 1cm 3cm 1cm},clip]{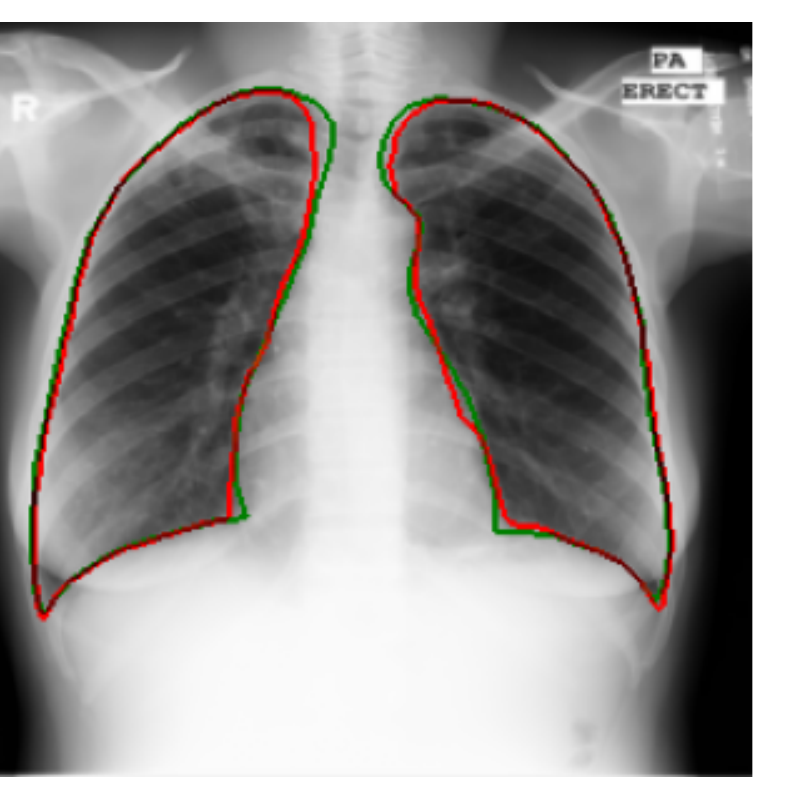}
    \\
    \noalign{\smallskip}
    {\Large\rotatebox{90}{MultiMix}} 
    &
    \includegraphics[width=0.2\linewidth, trim={4cm 1cm 3cm 1cm},clip]{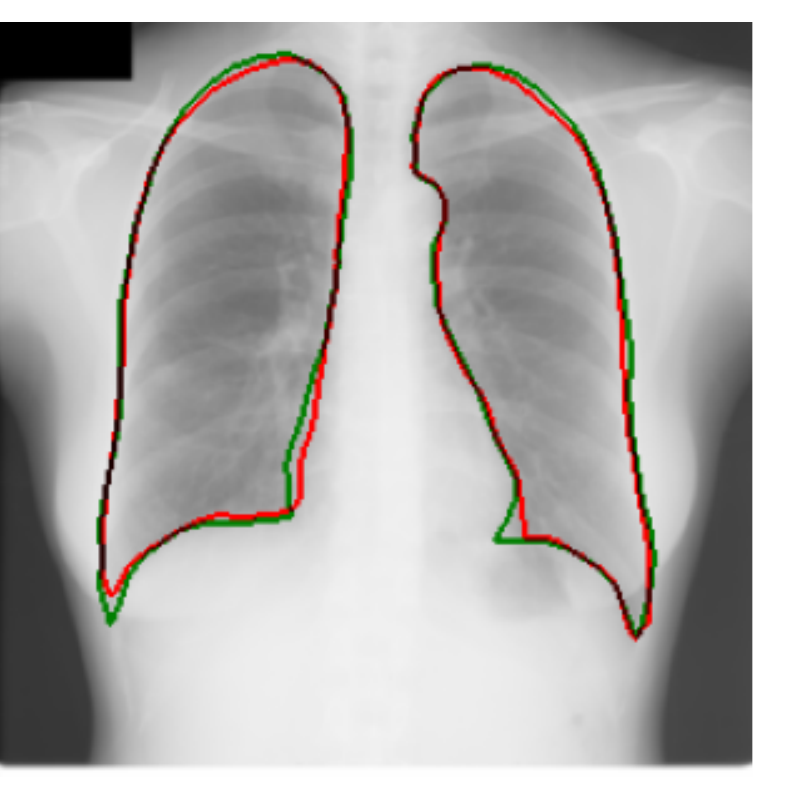}
    &
    \includegraphics[width=0.2\linewidth, trim={4cm 1cm 3cm 1cm},clip]{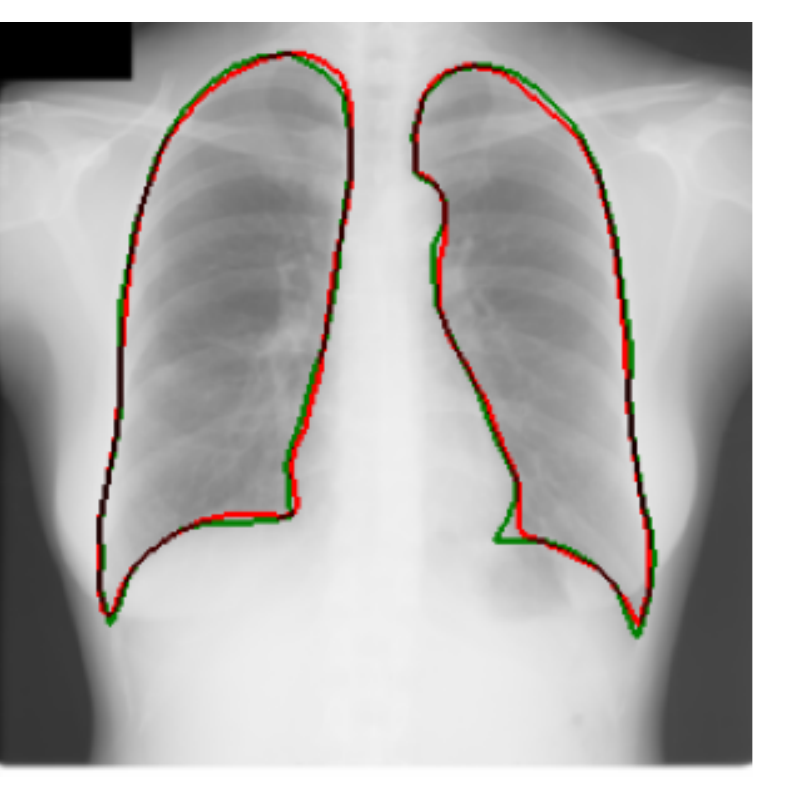}
    &
    \includegraphics[width=0.2\linewidth, trim={4cm 1cm 3cm 1cm},clip]{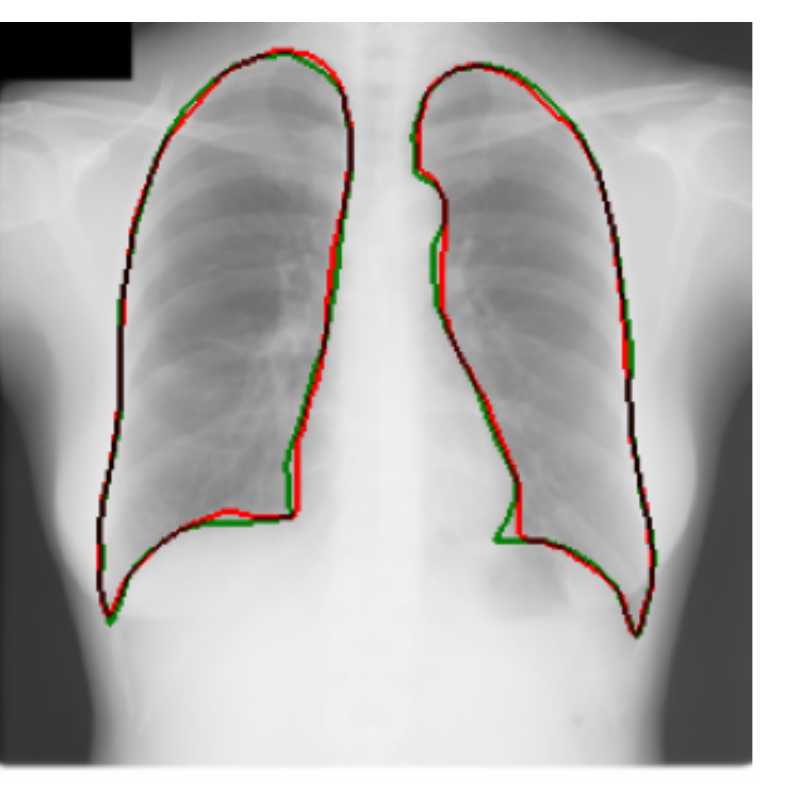}
    &
    \includegraphics[width=0.2\linewidth, trim={4cm 1cm 3cm 1cm},clip]{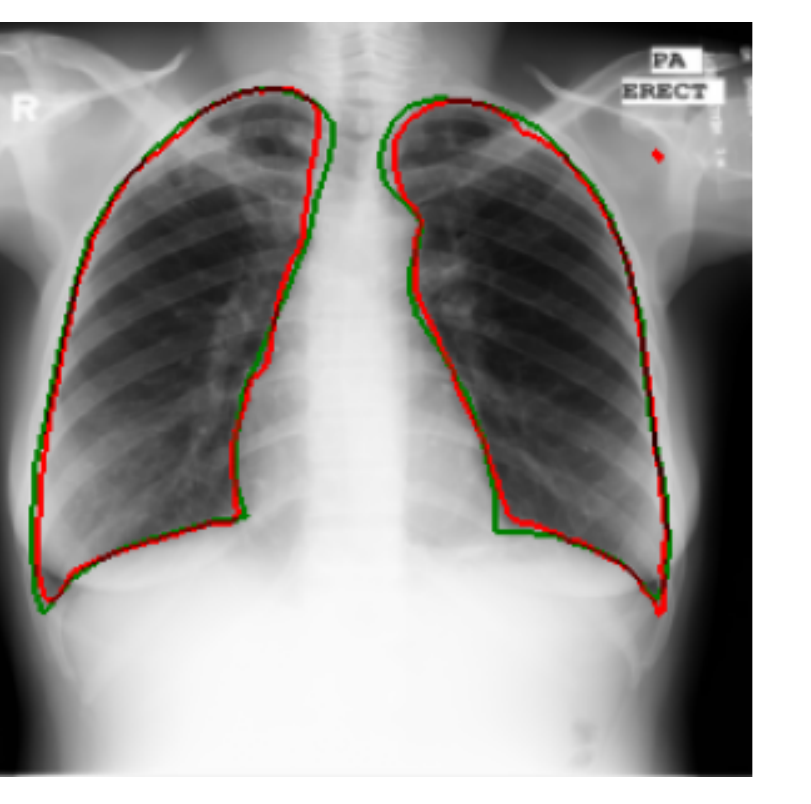}
    &
    \includegraphics[width=0.2\linewidth, trim={4cm 1cm 3cm 1cm},clip]{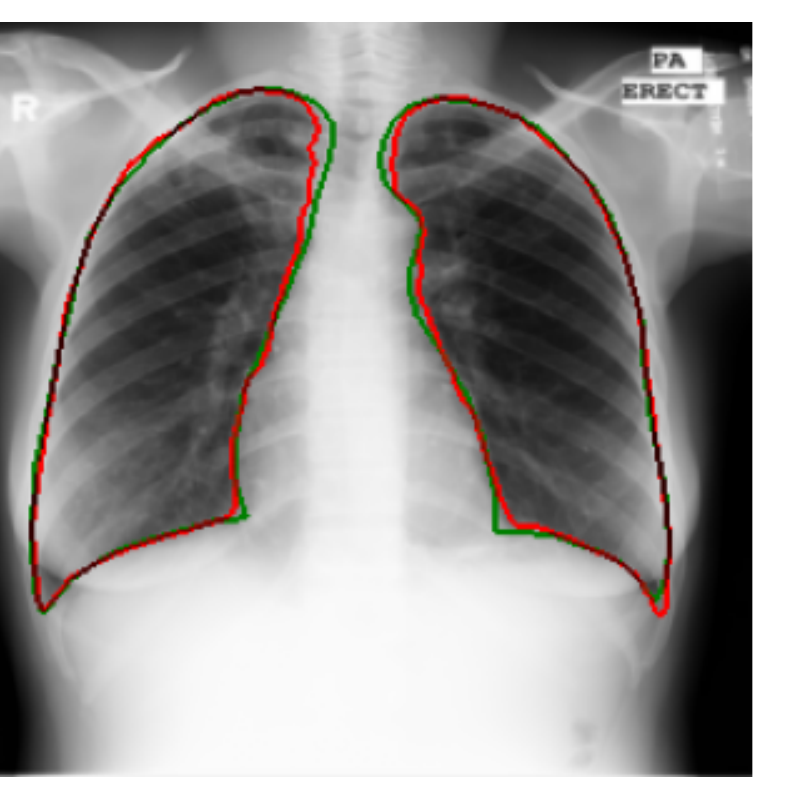}
    &
    \includegraphics[width=0.2\linewidth, trim={4cm 1cm 3cm 1cm},clip]{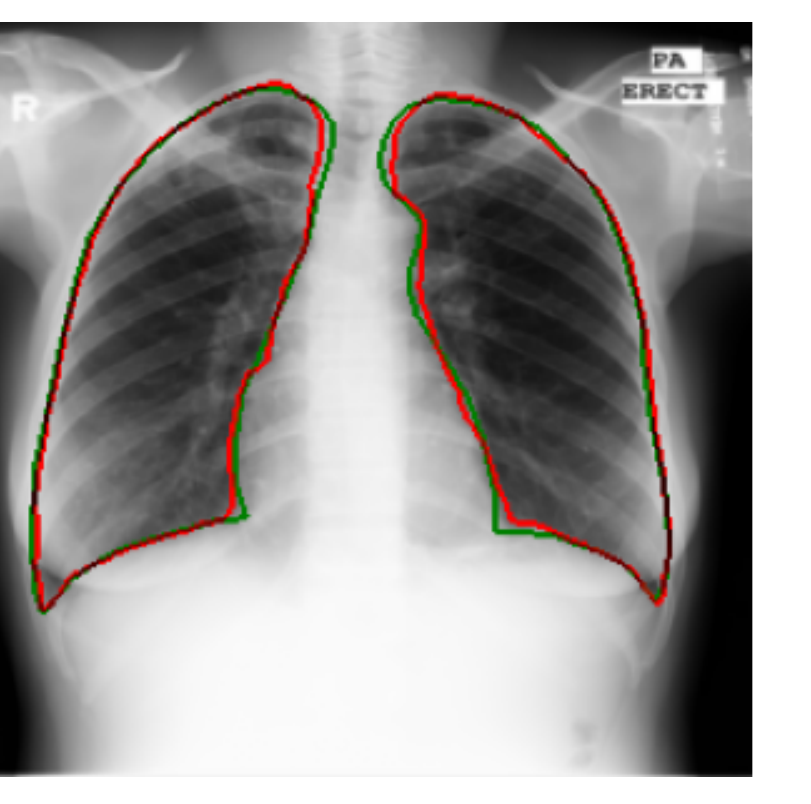}
    \\
    \end{tabular} 
    }
    \caption{Boundary visualization of the predicted segmentations in a chest X-ray depicts the superiority of MultiMix. Color code: green (reference), red (predicted).}
    \label{fig:lung_vis}
\end{figure}

\subsection{Results and Discussion}

%Table: In-domain
 \begin{table*}[t]
\centering
\setlength{\tabcolsep}{4pt}
\caption{Classification and segmentation performance comparison with varying label proportions. Best fully-supervised scores are underlined and best semi-supervised scores are bolded.}
\medskip
\label{table:scores}
\resizebox{0.821\linewidth}{!}{
\begin{tabular}{@{} lc cc c ccc c ccc c ccc c ccc@{}}
            \toprule
          \multirow{3}{*}{Model}
           &
           \phantom{a}
           &
           \multirow{3}{*}{$|\mathcal{D}^c_l|$}
           &
           \multirow{3}{*}{$|\mathcal{D}^s_l|$}
           &
           \phantom{a}
           &
           \multicolumn{7}{c}{In-Domain}
           &
           \phantom{a}
           &
           \multicolumn{7}{c}{Cross-Domain}
           \\
           \cmidrule{6-12}\cmidrule{14-20}
           && &&& 
           \multicolumn{3}{c}{Classification}
           &&
          \multicolumn{3}{c}{Segmentation}
          &&
          \multicolumn{3}{c}{Classification}
           &&
          \multicolumn{3}{c}{Segmentation}
          \\
          \cmidrule{6-8}\cmidrule{10-12}\cmidrule{14-16}\cmidrule{18-20}
          &&&&& Acc & F1-N & F1-P && DS & HD & SSIM && 
          Acc & F1-N & F1-P && DS & HD & SSIM \\
          \midrule
          \multirow{3}{*}{\rotatebox{45}{U-Net}} 
           &&
           --- & 10
           &&
           --- & --- & --- && 0.634 & 2.899 & 0.810 %seg-sample
           &&
           --- & --- & --- && 0.555 & 8.691 & 0.680 %seg-sample
           \\
           &&
           --- & 50
           &&
           --- & --- & --- && 0.855 & 0.341 & 0.904 %seg-sample
           &&
           --- & --- & --- && 0.763 & 2.895 & 0.870 %seg-sample
           \\
           &&
           --- & Full
           &&
           --- & --- & --- && 0.915 & 0.104 & 0.929 %seg-sample
           &&
           --- & --- & --- && 0.838 & 1.414 & 0.929 %seg-sample
           \\
           \midrule
           \multirow{3}{*}{\rotatebox{45}{Enc}} 
           &&
           100 & ---
           &&
           0.732 & 0.424 & 0.806 && --- & --- & --- %seg-sample
           &&
           0.352 & 0.070 & 0.506 && --- & --- & --- %seg-sample
           \\
           &&
           1000 & ---
           &&
           0.773 & 0.546 & 0.842 && --- & --- & --- %seg-sample
           &&
           0.390 & 0.192 & 0.508 && --- & --- & --- %seg-sample
           \\
           &&
           Full & ---
           &&
           0.737 & 0.534 & 0.838 && --- & --- & --- %seg-sample
           &&
           0.434 & 0.296 & 0.524 && --- & --- & --- %seg-sample
           \\
           \midrule
          \multirow{3}{*}{\rotatebox{45}{EncSSL}}
           &&
           100 & ---
           &&
           0.780 & 0.570 & 0.844 && --- & --- & --- %seg-sample
           &&
           0.402 & 0.222 & 0.510 && --- & --- & --- %seg-sample
           \\
           &&
           1000 & ---
           &&
           0.822 & 0.692 & 0.876 && --- & --- & --- %seg-sample
           &&
           0.486 & 0.380 & 0.530 && --- & --- & --- %seg-sample
           \\
           &&
           Full & ---
           &&
           0.817 & 0.680 & 0.872 && --- & --- & --- %seg-sample
           &&
           0.510 & 0.472 & 0.538 && --- & --- & --- %seg-sample
           \\
           \midrule
           \multirow{9}{*}{\rotatebox{45}{UMTL}}
           &&
           100 & 10
           &&
           0.707 & 0.443 & 0.797 && 0.626 & 4.323 & 0.908 %seg-sample
           &&
           0.350 & 0.045 & 0.510 && 0.586 & 7.156 & 0.836 %seg-sample
           \\
           &&
           100 & 50
           &&
           0.655 & 0.683 & 0.853 && 0.647 & 4.733 & 0.881 %seg-sample
           &&
           0.363 & 0.085 & 0.515 && 0.580 & 7.013 & 0.825 %seg-sample
           \\
           &&
           100 & Full
           &&
           0.706 & 0.416 & 0.804 && 0.696 & 3.908 & 0.911 %seg-sample
           &&
           0.342 & 0.015 & 0.508 && 0.607 & 6.398 & 0.863 %seg-sample
           \\
           &&
           1000 & 10
           &&
           0.750 & 0.490 & 0.825 && 0.761 & 3.050 & 0.926 %seg-sample
           &&
           0.413 & 0.263 & 0.507 && 0.676 & 3.268 & 0.833 %seg-sample
           \\
           &&
           1000 & 50
           &&
           0.749 & 0.510 & 0.833 && 0.768 & 2.606 & 0.938 %seg-sample
           &&
           0.400 & 0.203 & 0.513 && 0.704 & 3.232 & 0.896 %seg-sample
           \\
           &&
           1000 & Full
           &&
           0.747 & 0.530 & 0.840 && 0.759 & 2.955 & 0.930 %seg-sample
           &&
           0.430 & 0.293 & 0.517 && 0.638 & 3.893 & 0.890 %seg-sample
           \\
           &&
           Full & 10
           &&
           0.744 & 0.515 & 0.828 && 0.909 & 0.903 & 0.521 %seg-sample
           &&
           0.455 & 0.365 & 0.525 && 0.737 & 0.917 & 0.879 %seg-sample
           \\
           &&
           Full & 50
           &&
           0.738 & 0.438 & 0.820 && 0.930 & 0.444 & 0.954 %seg-sample
           &&
           0.444 & 0.332 & 0.522 && 0.868 & 0.742 & 0.894 %seg-sample
           \\
          &&
          Full & Full
          &&
          0.731 & 0.447 & 0.822 && 0.932 & 0.372 & 0.957 %seg-sample
          &&
          0.443 & 0.328 & 0.520 && 0.854 & 0.866 & 0.792 %seg-sample
           \\
           \midrule
           \multirow{9}{*}{\rotatebox{45}{UMTLS}}
           &&
           100 & 10
           &&
           0.704 & 0.358 & 0.806 && 0.922 & 4.005 & 0.891 %seg-sample
           &&
           0.344 & 0.006 & 0.510 && 0.797 & 5.754 & 0.807 %seg-sample
           \\
           &&
           100 & 50
           &&
           0.701 & 0.336 & 0.796 && 0.926 & 4.393 & 0.894 %seg-sample
           &&
           0.364 & 0.098 & 0.506 && 0.828 & 6.412 & 0.826 %seg-sample
           \\
           &&
           100 & Full
           &&
           0.713 & 0.442 & 0.794 && 0.931 & 3.983 & 0.920 %seg-sample
           &&
           0.342 & 0.008 & 0.510 && 0.838 & 6.321 & 0.834 %seg-sample
           \\
           &&
           1000 & 10
           &&
           0.740 & 0.482 & 0.828 && 0.948 & 2.546 & 0.924 %seg-sample
           &&
           0.378 & 0.138 & 0.512 && 0.844 & 3.921 & 0.854 %seg-sample
           \\
           &&
           1000 & 50
           &&
           0.771 & 0.566 & 0.844 && 0.965 & 2.083 & 0.941 %seg-sample
           &&
           0.392 & 0.186 & 0.514 && 0.883 & 3.017 & 0.888 %seg-sample
           \\
           &&
           1000 & Full
           &&
           0.742 & 0.497 & 0.830 && 0.962 & 1.758 & 0.935 %seg-sample
           &&
           0.370 & 0.130 & 0.510 && 0.898 & 4.150 & 0.905 %seg-sample
           \\
           &&
           Full & 10
           &&
           0.747 & 0.500 & 0.830 && 0.955 & \textbf{0.268} & 0.936 %seg-sample
           &&
           0.470 & 0.398 & 0.524 && 0.881 & 0.862 & 0.888 %seg-sample
           \\
           &&
           Full & 50
           &&
           0.737 & 0.433 & 0.820 && 0.972 & 0.560 & 0.953 %seg-sample
           &&
           0.413 & 0.270 & 0.510 && 0.917 & 0.658 & 0.919 %seg-sample
           \\
           &&
           Full & Full
           &&
           0.723 & 0.413 & 0.817 && 0.974 & \underline{0.327} & 0.957 %seg-sample
           &&
           0.433 & 0.315 & 0.513 && 0.916 & 0.882 & 0.921 %seg-sample
           \\
           \midrule
           \multirow{9}{*}{\rotatebox{45}{MultiMix}}
           &&
           100 & 10
           &&
           0.800 & 0.594 & 0.856 && 0.954 & 0.695 & 0.938 %seg-sample
           &&
           0.440 & 0.164 & 0.510 && 0.857 & 1.227 & 0.863 %seg-sample
           \\
           &&
           100 & 50
           &&
           0.824 & 0.613 & 0.854 && 0.971 & 0.681 & 0.951 %seg-sample
           &&
           0.370 & 0.036 & 0.510 && 0.889 & 1.061 & 0.890 %seg-sample
           \\
           &&
           100 & Full
           &&
           0.792 & 0.593 & 0.854 && 0.973 & 0.636 & 0.954 %seg-sample
           &&
           0.500 & 0.300 & 0.510 && 0.899 & 0.647 & 0.906 %seg-sample
           \\
           &&
           1000 & 10
           &&
           0.817 & 0.647 & 0.865 && 0.954 & 0.902 & 0.932 %seg-sample
           &&
           0.520 & 0.386 & 0.530 && 0.862 & 1.307 & 0.878 %seg-sample
           \\
           &&
           1000 & 50
           &&
           0.825 & 0.650 & 0.860 && 0.970 & 0.811 & 0.950 %seg-sample
           &&
           0.540 & 0.500 & 0.536 && 0.912 & 1.293 & 0.907 %seg-sample
           \\
           &&
           1000 & Full
           &&
           0.830 & 0.586 & 0.856 && \textbf{0.974} & 0.643 & 0.953 %seg-sample
           &&
           \textbf{0.570} & \textbf{0.620} & 0.510 && \textbf{0.936} & 0.803 & \textbf{0.932} %seg-sample
           \\
           &&
           Full & 10
           &&
           0.840 & 0.730 & 0.880 && 0.954 & 0.621 & 0.935 %seg-sample
           &&
           0.550 & 0.430 & 0.534 && 0.886 & 0.746 & 0.894 %seg-sample
           \\
           &&
           Full & 50
           &&
           \textbf{0.854} & \textbf{0.760} & \textbf{0.890} && 0.972 & 0.692 & \textbf{0.956} %seg-sample
           &&
           0.560 & 0.570 & \textbf{0.550} && 0.935 & \textbf{0.515} & 0.930 %seg-sample
           \\
           &&
           Full & Full
           &&
           \underline{0.843} & \underline{0.740} & \underline{0.890} && \underline{0.975} & 0.528 & \underline{0.960} %seg-sample
           &&
           \underline{0.520} & \underline{0.490} & \underline{0.550} && \underline{0.943} & \underline{0.417} & \underline{0.937} %seg-sample
           \\
           \bottomrule
          \end{tabular}}
          \end{table*}

\textbf{In-Domain:}
Our models were trained on the CheX and JSRT datasets. As revealed by the in-domain results (Table~\ref{table:scores}), model performance is improved with the subsequent inclusion of each of the novel components on the backbone network. For classification, our semi-supervised algorithm has significantly improved performance compared to the baseline models, as with $\min|\mathcal{D}^c_l|$, and it even outperforms the fully-supervised baseline. 
For segmentation, the saliency bridge, our primary addition, yields large improvements over the baseline U-Net and U-MTL. Again, with $\min|\mathcal{D}^s_l|$, a 30\% performance gain over its counterparts proves the effectiveness of our MultiMix model. Fig.~\ref{fig:consistency} depicts better consistency by our proposed model over the baselines. For fair and proper comparison, we use the same backbone U-Net and the same classification branch for all the models. The segmented lung boundary visualizations also show good agreement with the reference masks by MultiMix over other models (Fig.~\ref{fig:lung_vis}).

\textbf{Cross-Domain:}
We validated our models on two cross-domain datasets, the NIHX and MCU datasets, for classification and segmentation respectively. The cross-domain results (Table~\ref{table:scores}) are as promising as the in-domain ones. The classification accuracy is increased with the introduction of MultiMix models. Due to the significant differences in the NIHX and CheX datasets, the scores are not as good as the in-domain results. Yet our model performs better than the other models. For segmentation, our MultiMix model again achieves better scores in all different metrics, with improved consistency over the baselines (Fig.~\ref{fig:consistency}).

\section{Conclusions}

We have presented a novel sparingly supervised, multitask learning model (MultiMix) for jointly learning classification and segmentation tasks. Through the incorporation of consistency augmentation and a novel saliency bridge module, MultiMix performs improved and consistent pneumonia detection and lung segmentation when trained on multi-source datasets, varying labeled data. Extensive experimentation using four different chest X-ray datasets demonstrated the effectiveness of MultiMix both in in-domain and cross-domain evaluations, for either tasks; indeed, outperforming a number of baselines. Our future work will focus on further improving MultiMix's cross-domain performance.

\balance
\bibliographystyle{IEEEbib}
\bibliography{refs}

\begin{thebibliography}{10}

\bibitem{chapelle2009semi}
Olivier Chapelle, Bernhard Scholkopf, and Alexander Zien,
\newblock ``Semi-supervised learning (chapelle, o. et al., eds.; 2006)[book
  reviews],''
\newblock {\em IEEE Transactions on Neural Networks}, vol. 20, no. 3, pp.
  542--542, 2009.

\bibitem{ruder2017overview}
Sebastian Ruder,
\newblock ``An overview of multi-task learning in deep neural networks,'' 2017.

\bibitem{imran2019semi}
Abdullah-Al-Zubaer Imran and Demetri Terzopoulos,
\newblock ``Semi-supervised multi-task learning with chest x-ray images,''
\newblock in {\em Machine Learning in Medical Imaging: 10th International
  Workshop, MLMI 2019, Held in Conjunction with MICCAI 2019, Shenzhen, China,
  October 13, 2019, Proceedings}. Springer Nature, 2019, vol. 11861, p. 151.

\bibitem{liu2008semi}
Qiuhua Liu, Xuejun Liao, and Lawrence Carin,
\newblock ``Semi-supervised multitask learning,''
\newblock in {\em Advances in Neural Information Processing Systems}, 2008, pp.
  937--944.

\bibitem{mehta2018ynet}
Sachin Mehta, Ezgi Mercan, Jamen Bartlett, Donald Weave, Joann~G. Elmore, and
  Linda Shapiro,
\newblock ``Y-net: Joint segmentation and classification for diagnosis of
  breast biopsy images,'' 2018.

\bibitem{simonyan2014deep}
Karen Simonyan, Andrea Vedaldi, and Andrew Zisserman,
\newblock ``Deep inside convolutional networks: Visualising image
  classification models and saliency maps,'' 2014.

\bibitem{ronneberger2015unet}
Olaf Ronneberger, Philipp Fischer, and Thomas Brox,
\newblock ``U-net: Convolutional networks for biomedical image segmentation,''
  2015.

\bibitem{sohn2020fixmatch}
Kihyuk Sohn, David Berthelot, Chun-Liang Li, Zizhao Zhang, Nicholas Carlini,
  Ekin~D. Cubuk, Alex Kurakin, Han Zhang, and Colin Raffel,
\newblock ``Fixmatch: Simplifying semi-supervised learning with consistency and
  confidence,'' 2020.

\bibitem{kermany2018identifying}
Daniel~S Kermany, Michael Goldbaum, Wenjia Cai, Carolina~CS Valentim, Huiying
  Liang, Sally~L Baxter, Alex McKeown, Ge~Yang, Xiaokang Wu, Fangbing Yan,
  et~al.,
\newblock ``Identifying medical diagnoses and treatable diseases by image-based
  deep learning,''
\newblock {\em Cell}, vol. 172, no. 5, pp. 1122--1131, 2018.

\bibitem{shiraishi2000development}
Junji Shiraishi, Shigehiko Katsuragawa, et~al.,
\newblock ``Development of a digital image database for chest radiographs with
  and without a lung nodule,''
\newblock {\em J of Roent}, 2000.

\bibitem{jaeger2014two}
Stefan Jaeger, Sema Candemir, et~al.,
\newblock ``Two public chest {X}-ray datasets for computer-aided screening of
  pulmonary diseases,''
\newblock {\em Quant Imag in Med and Surg}, 2014.

\bibitem{wang2017chestx}
Xiaosong Wang, Yifan Peng, Le~Lu, Zhiyong Lu, Mohammadhadi Bagheri, and
  Ronald~M Summers,
\newblock ``Chestx-ray8: Hospital-scale chest x-ray database and benchmarks on
  weakly-supervised classification and localization of common thorax
  diseases,''
\newblock in {\em Proceedings of the IEEE conference on computer vision and
  pattern recognition}, 2017, pp. 2097--2106.

\end{thebibliography}

\end{document}